\documentclass{article}

\usepackage{arxiv}

\usepackage[utf8]{inputenc} 
\usepackage[T1]{fontenc}    
\usepackage{hyperref}       
\usepackage{url}            
\usepackage{booktabs}       
\usepackage{amsfonts}       
\usepackage{nicefrac}       
\usepackage{microtype}      
\usepackage{lipsum}
\usepackage{graphicx}
\usepackage{amsmath}
\usepackage[authoryear]{natbib}
\graphicspath{ {./images/} }

\usepackage{amssymb,amsfonts}
\usepackage{multirow}
\usepackage{subcaption}

\title{DuaDeep-SeqAffinity: Dual-Branch Deep Learning \\ for Tri-Stream Sequence-Based Antibody--Antigen Affinity Prediction}

\author{
 Aicha Boutorh $\thanks{Corresponding author}$ \\
  National School of Artificial Intelligence (ENSIA)\\
  Sidi Abdallah Campus, Algiers, Algeria\\
  \texttt{aicha.boutorh@ensia.edu.dz} \\
   \And
 Soumia Bouyahiaoui \\
  National School of Artificial Intelligence (ENSIA)\\
  Sidi Abdallah Campus, Algiers, Algeria\\
\texttt{soumia.bouyahiaoui@ensia.edu.dz} \\
\And
 Manel Kara Laouar \\
  National School of Artificial Intelligence (ENSIA) \\
  Sidi Abdallah Campus, Algiers, Algeria\\
  \texttt{manel.karalaouar@ensia.edu.dz}\\
  \And
 Sara Belhadj \\
  National School of Artificial Intelligence (ENSIA) \\
  Sidi Abdallah Campus, Algiers, Algeria\\
  \texttt{sara.belhadj@ensia.edu.dz} \\
  \And
 Nour El Yakine Guendouz \\
  National School of Artificial Intelligence (ENSIA) \\
  Sidi Abdallah Campus, Algiers, Algeria\\
  \texttt{nour-el-yakine.guendouz@ensia.edu.dz} \\
  \And
 Asma Boutorh \\
  Lamine Debaghine University Hospital Center, CHU Bab El Oued, \\
  Algiers, Algeria\\
  \texttt{bout.asma.ph@gmail.com} \\
}

\begin{document}
\maketitle
\begin{abstract}
\textbf{DuaDeep-SeqAffinity} is a sequence-only deep learning framework
that predicts antibody--antigen binding affinity directly from primary
amino acid sequences, avoiding the cost and scarcity of resolved
three-dimensional structures. The antigen and the antibody heavy and
light chains are processed as three independent streams, each embedded
with a frozen ESM-2 protein language model and passed through parallel
Transformer and convolutional neural network (CNN) branches before late
fusion, a decoupled design intended to preserve local
complementarity-determining region (CDR) signal that monolithic
encoders can dilute. On a sequence-disjoint split of the AbRank
benchmark, the model achieves a Pearson correlation of $0.683 \pm
0.005$, an $R^2$ of $0.460 \pm 0.007$, and a pairwise ranking AUC of
$0.895 \pm 0.004$, significantly
outperforming single-branch ablations (paired $t$-test, $p < 0.05$). Attention-map and gradient-based saliency analyses further show
that the model preferentially attends to CDR loops and candidate
epitope residues, supporting its use as a scalable, structure-free tool
for high-throughput antibody screening. \end{abstract}



\section{Introduction}
\label{Intro}

The high-affinity interaction between antigens and antibodies is the
cornerstone of the adaptive immune response and a primary objective of
therapeutic intervention. Accurate prediction of binding affinity
($\mathrm{p}K_d$) is essential for identifying high-potency leads,
enabling rational vaccine design, and engineering sensitive diagnostic
platforms.\cite{lu2020development,tang2023machine} As drug discovery
shifts toward high-throughput computational screening, developing
scalable predictive models that bypass the prohibitive cost of
experimental structural determination has become a central goal of
computational chemistry and cheminformatics, paralleling the broader
shift toward data-driven structure--activity models in small-molecule
drug design. Traditional computational methods for protein--protein
interactions (PPIs) rely heavily on structure-based
modeling.\cite{kozakov2017cluspro,gaudreault2025ai} However, these
approaches are constrained by the scarcity of high-resolution
three-dimensional complexes in the Protein Data Bank
(PDB).\cite{berman2000protein} Deep-learning advances such as
AlphaFold3 have revolutionized general protein-complex modeling, yet
they continue to struggle with the inherent conformational flexibility
of antibody complementarity-determining regions (CDRs); recent
benchmarks indicate persistent limitations in modeling these interfaces
under realistic docking
conditions.\cite{abramson2024accurate,hitawala2025does} Even
specialized, faster antibody structure predictors that avoid full
physics-based docking, such as IgFold,\cite{ruffolo2023igfold} still
require an explicit structure-generation step before an affinity or
interface score can be computed, adding latency and an additional
source of error propagation for high-throughput screening pipelines.
This ``structural bottleneck'' motivates a shift toward sequence-only
models that extract biophysical signatures directly from primary amino
acid sequences.\cite{jin2024attabseq} Large-scale protein language
models (PLMs) pretrained on evolutionary data, from ESM-2 to the more
recent multimodal ESM3,\cite{lin2023evolutionary,hayes2025esm3} encode
rich structural and functional information, including zero-shot
sensitivity to the effects of mutations on protein
function,\cite{meier2021zeroshot} making them an attractive substrate
for structure-free affinity prediction.

However, sequence-based modeling of antibodies presents a unique
``scaling challenge.'' The germline-encoded framework regions, which
provide structural stability, often dominate the raw sequence signal,
potentially diluting the high-frequency, localized variation within the
CDR loops that dictates binding specificity. Current models often
utilize monolithic backbones that merge these multi-scale signals
prematurely, leading to a loss of fine-grained information.

In this work, we introduce \textit{DuaDeep-SeqAffinity}, a framework
employing a \textit{tri-stream, dual-branch architecture} designed to
resolve these multi-scale dependencies. Our core contribution is the
enforcement of an architectural ``separation of concerns'': we
explicitly segregate the learning of \textit{global evolutionary
grammar} from \textit{fine-grained local binding motifs}. By utilizing
parallel branches -- Transformer encoders for long-range contextual
dependencies and 1D-CNNs for local physicochemical signatures -- we aim
to prevent the dilution of paratope-specific signals. This allows
\textit{DuaDeep-SeqAffinity} to recover latent, structurally relevant
sequence dependencies and achieve state-of-the-art sequence-only
performance without three-dimensional templates.

Our contributions are four-fold: (i) we propose a tri-stream,
dual-branch architecture specifically tuned to the hierarchical nature
of the antibody--antigen interface; (ii) we demonstrate, with paired
statistical testing across four random seeds, that architectural
decoupling significantly outperforms monolithic single-branch ablations
in affinity regression and ranking tasks; (iii) we provide mechanistic
interpretability through attention and gradient-based saliency mapping,
and explicitly examine the scope and limits of this evidence; and (iv)
we provide a  discussion  that should inform how these results, and those of
related sequence-only affinity predictors, are interpreted.

The remainder of this paper is organized as follows: \textit{Related
Work} reviews relevant literature and existing methodologies;
\textit{Methods} details the proposed DuaDeep-SeqAffinity architecture;
\textit{Results and Discussion} presents the experimental results,
statistical analysis, interpretability findings, and a critical
discussion of limitations; and \textit{Conclusions} summarizes our
findings and outlines future research directions.


\section{Background and Related Work}
\label{Related}
 
\subsection{Antibody--Antigen Interaction}
 
\begin{figure}[t]
    \centering
    \includegraphics[width=0.75\linewidth]{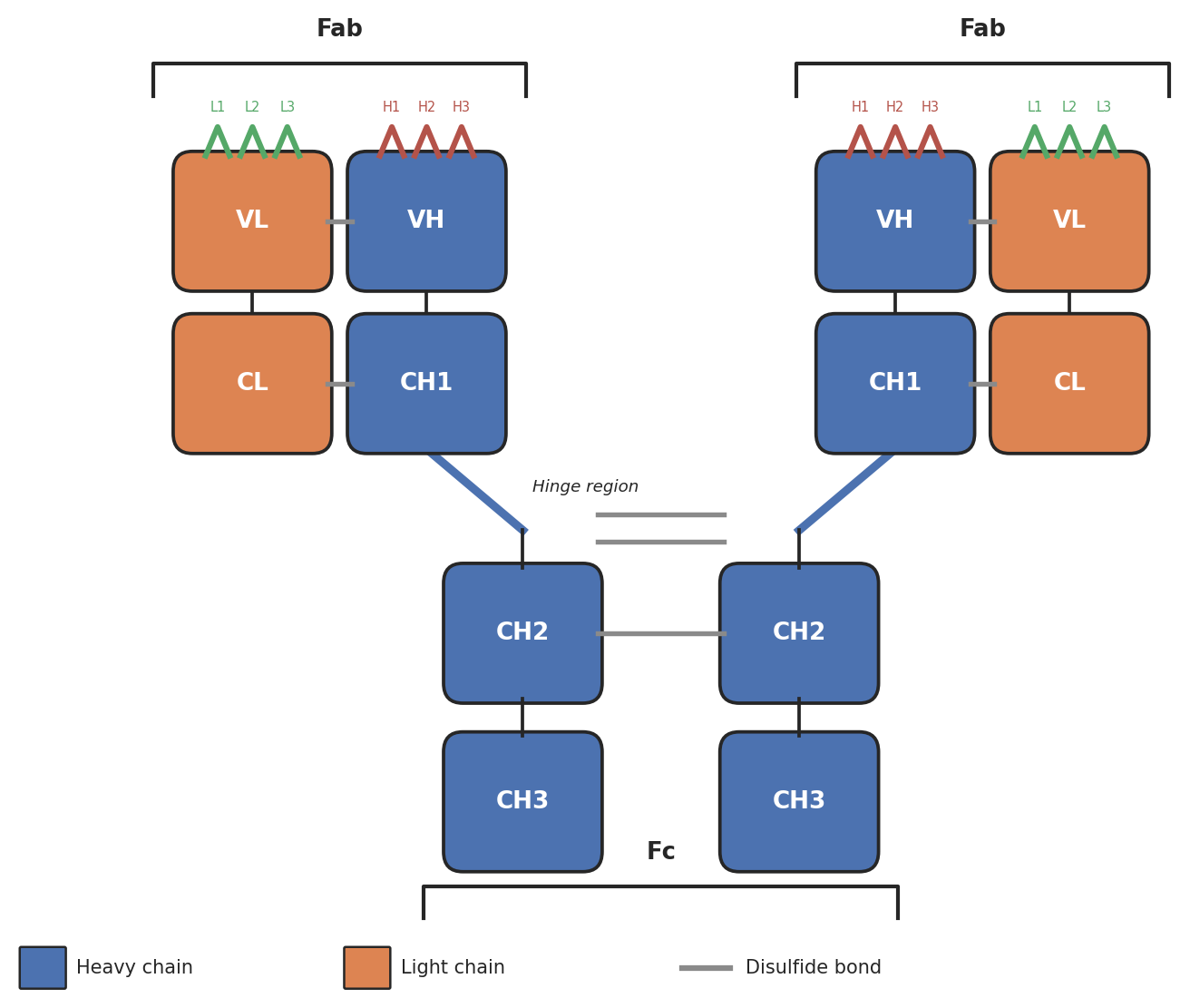}
    \caption{Overall antibody (IgG) structure.\cite{zhu202450} Two heavy
    chains and two light chains are held together by disulfide bonds
    (grey). Each chain contributes a variable domain (VH, VL) and one or
    more constant domains (CH1--CH3, CL). The variable domains at the
    tip of each Fab arm carry three complementarity-determining regions
    per chain (H1--H3, L1--L3), which together form the paratope, the
    antigen-binding site. The two Fc-region constant domains (CH2--CH3)
    mediate effector functions and are not directly involved in antigen
    recognition.}
    \label{fig:antibodies}
\end{figure}

The antibody-mediated immune response is a critical component of the
adaptive immune system, responsible for the highly specific recognition
and neutralization of foreign substances, including viruses, bacteria,
and toxins. Antibodies, or immunoglobulins (Ig), are Y-shaped
glycoproteins synthesized by B
lymphocytes.\cite{chiu2019antibody,oostindie2022avidity,padlan1977structural}
Each antibody identifies a specific molecular region on an antigen,
known as an epitope, via its paratope, the physical binding site
located at the tips of the antibody's variable regions.\cite{zhu202450}
The architecture of these variable regions is defined by six
hypervariable loops termed complementarity-determining regions (CDRs),
with three situated on the heavy chain (H1, H2, H3) and three on the
light chain (L1, L2, L3). These loops collectively dictate the
paratope's shape and physicochemical complementarity to the antigen,
thereby determining binding affinity (Figure~\ref{fig:antibodies}).
Among these, the CDR-H3 loop is characterized by the highest degree of
structural and sequence diversity, often playing a dominant role in
antigen recognition.\cite{chiu2019antibody} The immense diversity of the
CDR repertoire arises from somatic recombination and affinity
maturation through hypermutation during B-cell development, enabling
the immune system to target an almost limitless array of antigenic
determinants.\cite{odegard2006targeting}
 
\subsection{Therapeutic Importance of Antibody Binding}
 
Over the last few decades, monoclonal antibodies (mAbs) have become a
central component of modern biopharmaceuticals. They are engineered to
bind specifically to defined molecular targets, allowing precise
modulation of biological pathways. Since the approval of the first
therapeutic antibody in 1986 (muromonab-CD3), the number of clinically
approved mAbs has grown steadily, with applications spanning oncology,
autoimmune disorders, and infectious
disease.\cite{singh2018monoclonal,bayer2019overview} Their success
relies on the ability to design antibodies with high specificity and
controlled effector functions, reducing off-target effects and enabling
targeted interventions. Additionally, mAbs serve as the basis for
advanced biologics, such as antibody--drug conjugates (ADCs) and
bispecific antibodies.\cite{lebasle2020physicochemical}
 
A recent and impactful example is the deployment of monoclonal
antibodies against SARS-CoV-2, the virus responsible for COVID-19.
Antibodies such as bebtelovimab, tixagevimab, and cilgavimab received
emergency use authorization for treatment and pre-exposure prophylaxis,
particularly benefiting patients unable to mount adequate immune
responses to vaccines. These cases demonstrate how antibody
therapeutics can complement vaccines and small-molecule drugs in
combating emerging
pathogens.\cite{widyasari2023review,hwang2022monoclonal,taylor2021neutralizing}
Beyond traditional discovery, the integration of deep learning has
catalyzed computational drug
repurposing\cite{mohanty2020application,zhou2020artificial,boutorh2022graph}
and generative antibody
design,\cite{callaway2023generative,huo2024deep,kong2022conditional,tan2025dyab}
enabling the rapid identification of neutralizing candidates from
existing biologics libraries, or the de novo generation of new ones, to
keep pace with rapid viral evolution. Given their immense clinical and
commercial significance, the development of high-throughput tools for
predicting antibody--antigen binding affinity has become a critical
pillar for rational drug design and pandemic preparedness.
 
\subsection{Experimental Methods for Measuring Binding Affinity}
 
Accurate quantification of antibody--antigen binding affinity is
essential for characterizing immune interactions and developing
therapeutic antibodies. Several experimental techniques have been
established, each offering different levels of precision, throughput,
and cost-effectiveness.
 
Radioimmunoassay (RIA) is one of the earliest quantitative methods,
using radiolabeled antigens or antibodies to detect and measure complex
formation. While RIA provides high sensitivity, it requires radioactive
materials, making it hazardous, expensive, and less favorable in modern
laboratories.\cite{block1979overview,vallejo1979radioimmunoassay} The
enzyme-linked immunosorbent assay (ELISA) employs enzyme-linked
antibodies that produce measurable colorimetric or fluorescent signals
upon antigen binding. ELISA is well suited to medium-throughput
screening but may have limited sensitivity for very weak or transient
interactions.\cite{mir2019immunoglobulins,tabatabaei2022elisa}
 
Surface plasmon resonance (SPR) provides real-time, label-free
measurement of molecular interactions, allowing determination of
kinetic parameters in addition to overall affinity constants. Despite
its precision, SPR requires specialized equipment and can be costly and
low-throughput for large-scale screening.\cite{sparks2018use} Bio-layer
interferometry (BLI) offers a more scalable alternative to SPR, though
with reduced sensitivity for small molecules or low-affinity
interactions.\cite{rhea2021determining}
 
Although these techniques yield accurate results, they are
resource-intensive, time-consuming, and not easily scalable for
high-throughput applications. The increasing availability of large
immunological datasets has enabled computational approaches that
predict binding affinity rapidly and at lower cost.
 
\subsection{Computational Approaches for Antibody--Antigen Affinity Prediction}
 
Computational methods provide a powerful alternative to experimental
affinity determination, enabling the rapid screening of expansive
molecular libraries. These approaches are broadly categorized into
structure-based and sequence-based paradigms.
 
\textbf{Structure-based methods} rely on the three-dimensional (3D)
coordinates of the antibody--antigen complex to estimate binding
energy. Traditional tools such as RosettaAntibody, FoldX, and CSM-AB
utilize physics-based force fields and geometric complementarity to
model the energetic properties of the binding
interface.\cite{weitzner2017modeling,myung2022csm} While these methods
offer high physical interpretability, they are inherently limited by
the ``structural bottleneck'': the scarcity and high cost of obtaining
high-resolution experimental structures via X-ray crystallography or
cryo-electron microscopy.\cite{zhu202450} Although recent deep learning
breakthroughs such as AlphaFold3 and Boltz-2 have significantly improved
complex structure prediction, they still encounter difficulty modeling
the high flexibility of the CDR-H3 loop and often require substantial
computational resources for large-scale affinity
ranking.\cite{abramson2024accurate,passaro2025boltz,ruffolo2023igfold}
 
\textbf{Sequence-based methods} circumvent structural requirements by
predicting affinity directly from amino acid sequences. Early models
primarily used hand-crafted physicochemical features (e.g., hydropathy,
polarity) coupled with traditional machine learning. Recent
advancements in protein language models (PLMs), including ESM-2,
ProtBERT, and AbLang, have since revolutionized this field by
providing high-dimensional, contextualized embeddings that implicitly
encode evolutionary and biophysical
information.\cite{lin2023evolutionary,elnaggar2021prottrans,olsen2022ablang}
Building on this, hybrid architectures that couple such embeddings with
task-specific sequence encoders,  for example, combining ESM-2 with a
bidirectional long short-term memory (Bi-LSTM) network to localize the
antibody--antigen binding interface,  have further improved
residue-level interaction prediction from sequence alone.\cite{li2025enhanced}
 
More recently, deep learning has shown superior performance by
capturing complex, non-linear relationships between sequence and
affinity. Early architectures such as PIPR, a Siamese residual
recurrent-convolutional network for general protein--protein
interaction prediction, and related recurrent/convolutional frameworks
established that end-to-end sequence encoders could match or exceed
hand-crafted-feature
baselines.\cite{chen2019multifaceted,lee2023recent} DG-Affinity, a
sequence-based deep learning framework designed specifically for
antibody--antigen affinity, demonstrated that combining PLM embeddings
with a ConvNeXt-style architecture can achieve Pearson correlations
exceeding 0.65 without any structural input; the model leverages two
specialized pretrained PLMs to capture domain-specific sequence
representations, TAPE for antigens and AbLang for
antibodies.\cite{yuan2023dg} The AntiBinder framework similarly employs
bidirectional cross-attention to capture the intrinsic mechanisms of
antigen--antibody interaction, highlighting the effectiveness of
attention-based sequence modeling.\cite{zhang2024antibinder} In a
related setting, sequence-only prediction of \emph{changes} in binding
affinity upon mutation, rather than the absolute affinity value, 
has also been shown to be both accurate and interpretable, supporting
practical antibody engineering
workflows.\cite{liu2025sequence}
 
Another prominent example of sequence-only modeling is MVSF-AB, which
addresses the lack of structural data through a multi-view learning
strategy operating at two levels: a semantic level, where ProteinBERT
embeddings and CNNs capture global biochemical semantics, and a residue
level, where physicochemical descriptors from the AAindex database are
combined via matrix decomposition and multilayer perceptrons (MLPs) to
learn binding-site relationships.\cite{li2025mvsf} While MVSF-AB
represents a significant step forward in feature fusion, it relies on
static residue-level descriptors and simple MLP architectures to model
local patterns.
 
Hybrid embedding strategies have also emerged that combine
general-purpose PLMs such as ESM-2 with antibody-specific language
models such as AntiBERTy, for example in an ESM-2 + AntiBERTy ranking
configuration.\cite{lin2023evolutionary,ruffolo2022antibody} When
deployed in this ranking setup, dual-embedding models capture both
global protein-language patterns and specific paratope--epitope
dependencies, outperforming traditional single-model baselines in
prioritizing high-affinity binders.\cite{wang2025supervised} MINT
(Multimeric Interaction Transformer) represents a further step in this
direction, treating the antibody and antigen as an interdependent pair
rather than processing sequences in isolation; it has demonstrated
state-of-the-art performance in zero-shot binding-affinity ranking and
mutational-impact assessment across diverse protein--protein
interaction benchmarks.\cite{ullanat2025learning}
 
To bridge the gap between biophysical rigor and sequence-only
scalability, recent frameworks have adopted explicitly hybrid
structure--sequence approaches. A prominent example is WALLE-Affinity,
a graph neural network (GNN) that integrates structural graph
representations of the antibody--antigen complex with PLM
embeddings.\cite{liu2025abrank,liu2024asep} WALLE-Affinity was initially
formulated as a regression model to estimate real-valued binding
scores, leveraging GNNs to absorb structural inaccuracies and flexible
domain dynamics while using PLM embeddings for biochemical
context.\cite{liu2024asep} More recently, the model has been reframed
as a pairwise ranking framework within the AbRank benchmark; this
version learns relative binding preferences rather than absolute
values, substantially improving robustness to experimental noise and
heterogeneous assay conditions.\cite{liu2025abrank}
 
Beyond antibody--antigen-specific frameworks, general-purpose
architectures that separate global and local (or set-level and
quality-level) signal into parallel branches have proven effective more
broadly across bioinformatics and genomics. CAT-CPI and DeepPhosPPI, for
instance, combine CNN and Transformer branches for compound--protein
interaction and phosphorylation-effect prediction, respectively, though
both treat their input sequences as monolithic entities rather than
decomposing them by biological
chain.\cite{li2022cat,li2025deepphosppi} In an adjacent genomic domain,
VAMP-Net couples a permutation-invariant Set Attention Transformer with
a quality-aware 1D-CNN to jointly model epistatic variant interactions
and sequencing-confidence signals for \emph{Mycobacterium tuberculosis}
drug-resistance prediction, underscoring the broader effectiveness of
decoupled attention/convolution fusion for interpretable genomic and
proteomic classification tasks beyond the antibody--antigen
setting.\cite{boutorh2025vampnet}
 
The framework proposed in this research, \textit{DuaDeep-SeqAffinity},
advances antibody--antigen affinity prediction by addressing
limitations shared by these prior approaches. While hybrid models such
as WALLE-Affinity rely on scarce 3D structural templates, and
sequence-only models such as MINT use standard Transformers that
flatten the feature representation across the whole sequence, DuaDeep
employs a dual-stream hybrid architecture in which Transformer encoders
and 1D-CNNs run in parallel within each of three independent chain
streams (antigen, heavy, and light). This explicitly separates the
extraction of long-range global context from fine-grained local motifs
before the multiscale features are integrated through a dedicated
fusion module, providing a scalable, sequence-only solution that aims
to capture complex binding signatures without requiring structural
data.


\section{Methods}
\label{method}

\begin{figure}[h]
    \centering
    \includegraphics[width=\linewidth]{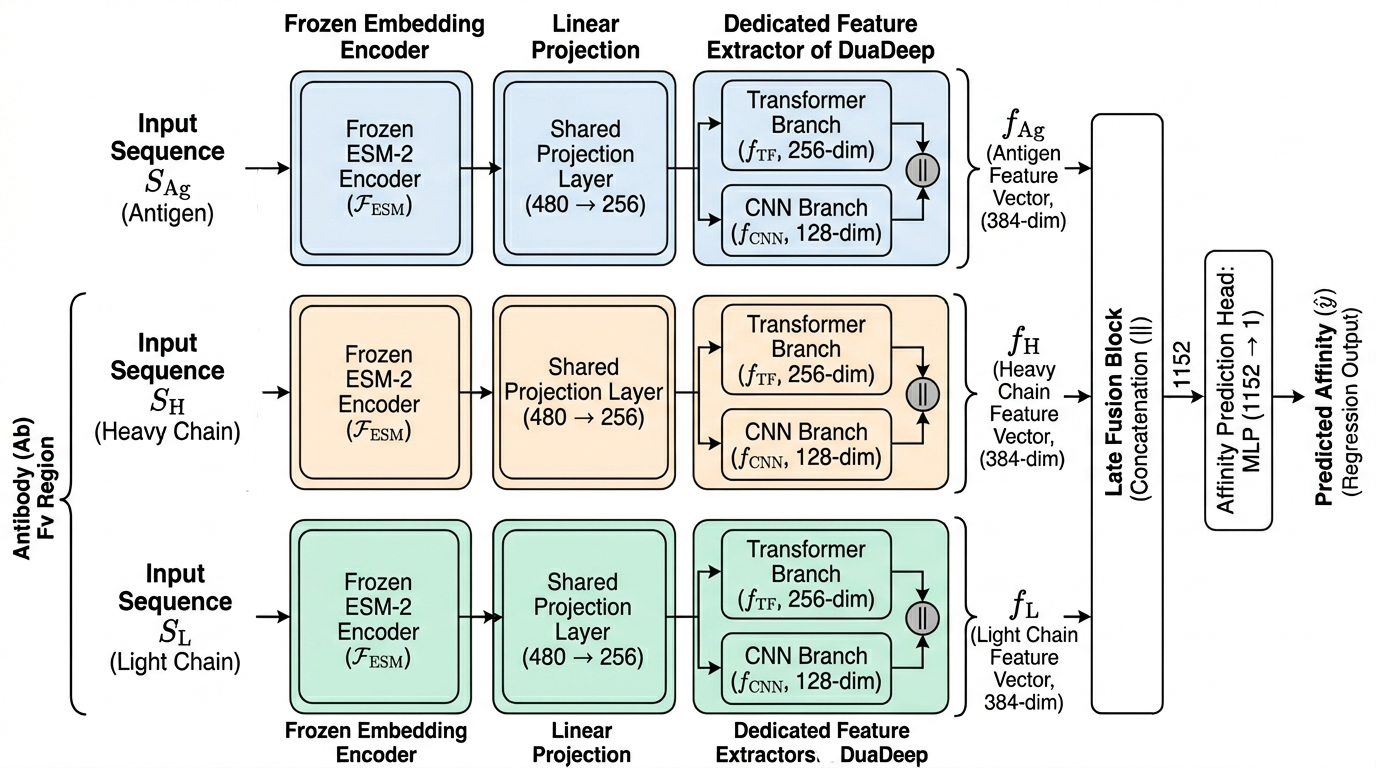}
    \caption{Dual-stream deep learning framework: architectural pipeline
    of DuaDeep-SeqAffinity. The framework processes the antigen
    ($S_{\mathrm{Ag}}$), heavy chain ($S_{\mathrm{H}}$), and light chain
    ($S_{\mathrm{L}}$) via three independent, structurally decoupled
    streams. Each stream utilizes a frozen ESM-2 backbone followed by a
    dedicated dual-branch (Transformer and CNN) feature extractor. A
    late-fusion strategy via vector concatenation ($\|$) preserves
    chain-specific representations prior to final affinity regression
    via a multilayer perceptron (MLP) head.}
    \label{fig:methodology}
\end{figure}

Our proposed framework, named \textbf{\textit{DuaDeep-SeqAffinity}}, is
designed to predict antigen--antibody affinity scores directly from raw
amino acid sequences, circumventing the need for computationally
expensive structure determination or prediction. This framework,
conceptually illustrated in Figure~\ref{fig:methodology}, is structured
as an end-to-end deep learning pipeline comprising three main,
interdependent stages: (1) sequence embedding using the pre-trained
ESM-2 protein language model; (2) dual-stream feature extraction,
combining Transformers and CNNs to capture global and local interaction
patterns for the antigen and antibody sequences separately; and (3) an
interaction prediction head for affinity scoring.

\subsection{ESM-2-Based Sequence Embedding Module}

The initial stage of our framework maps raw amino acid sequences into
high-dimensional, context-aware numerical manifolds. We use the ESM-2
protein language model for this purpose, to capture the evolutionary
grammar and biophysical constraints implicit in protein
sequences.\cite{lin2023evolutionary}

Let the antigen sequence be defined as $\mathbf{S}_{\mathrm{Ag}} = (a_1,
a_2, \dots, a_L)$ of length $L$. The antibody is represented by its
constituent heavy and light chains, denoted $\mathbf{S}_{\mathrm{H}} =
(h_1, h_2, \dots, h_{M_H})$ and $\mathbf{S}_{\mathrm{L}} = (l_1, l_2,
\dots, l_{M_L})$, where $M_H$ and $M_L$ are their respective lengths.
The sequences $\mathbf{S}_{\mathrm{Ag}}$, $\mathbf{S}_{\mathrm{H}}$, and
$\mathbf{S}_{\mathrm{L}}$ are processed independently through the
pre-trained ESM-2 encoder $\mathcal{F}_{\mathrm{ESM}}$ to generate
residue-level feature embeddings, where $D_E$ is the embedding
dimensionality (Equations \ref{eq:embed_ag}, \ref{eq:embed_h} and \ref{eq:embed_l} ):

\begin{align}
\mathbf{E}_{\mathrm{Ag}} &= \mathcal{F}_{\mathrm{ESM}}(\mathbf{S}_{\mathrm{Ag}}) \in \mathbb{R}^{L \times D_E}
 \label{eq:embed_ag}\\
\mathbf{E}_{\mathrm{H}}  &= \mathcal{F}_{\mathrm{ESM}}(\mathbf{S}_{\mathrm{H}})  \in \mathbb{R}^{M_H \times D_E} 
\label{eq:embed_h}\\
\mathbf{E}_{\mathrm{L}}  &= \mathcal{F}_{\mathrm{ESM}}(\mathbf{S}_{\mathrm{L}})  \in \mathbb{R}^{M_L \times D_E} 
\label{eq:embed_l}
\end{align}

For the reported experiments, $\mathcal{F}_{\mathrm{ESM}}$ is the
frozen \texttt{esm2\_t12\_35M\_UR50D} checkpoint ($D_E = 480$; see
Table~\ref{tab:hyperparams}). 

\subsection{Dual-Stream Feature Extraction: Hybrid Transformer--CNN}

The core of the \textit{DuaDeep} architecture is a multi-scale feature
extraction module designed to capture hierarchical dependencies within
the ESM-2 manifold. To preserve the structural integrity of the
paratope--epitope interface, the model operates across three
independent streams (antigen, heavy chain, and light chain), each
containing two parallel branches (Transformer and CNN).

Let $\mathbf{E}_X \in \mathbb{R}^{L_X \times D_E}$ denote the ESM-2
embeddings for a protein sequence $X$. Before feature extraction, we
apply a shared linear projection $\mathbf{W}_p \in \mathbb{R}^{D_E
\times d_m}$ to map the embeddings into a latent space of dimensionality
$d_m = 256$ (Equation \ref{eq:proj}):

\begin{equation}
\mathbf{Z}_X = \mathbf{E}_X \mathbf{W}_p \in \mathbb{R}^{L_X \times d_m}.
\label{eq:proj}
\end{equation}

The novelty of the \textit{DuaDeep} architecture lies in its decoupled
inductive bias. While existing frameworks such as
DG-Affinity\cite{yuan2023dg} utilize centralized backbones, our design
enforces a separation of representation scales:
\begin{itemize}
\item \textbf{Global Contextual Branch (Transformer):} models the
long-range evolutionary grammar and structural scaffolds via
self-attention mechanisms.
\item \textbf{Local Motif Branch (CNN):} uses 1D convolutions to
isolate local physicochemical neighborhoods and hypervariable patterns
within the CDR loops.
\end{itemize}
By maintaining these branches independently until late fusion, we aim
to mitigate the ``signal dilution'' effect observed in monolithic
models such as CAT-CPI,\cite{li2022cat} where fine-grained local
binding motifs are often overshadowed by dominant global sequence
patterns.

\subsubsection{Transformer Branch}

Each independent stream, antigen, heavy chain, and light chain,
processes its projected embeddings $\mathbf{Z}_X \in \mathbb{R}^{L_X
\times d_m}$ through a dedicated Transformer encoder that captures
long-range evolutionary dependencies and global contextual scaffolds
within each sequence. A single encoder layer consists of the following
components.

\textit{1. Multi-Head Self-Attention (MHSA).} We project $\mathbf{Z}_X$
into query ($\mathbf{Q}$), key ($\mathbf{K}$), and value ($\mathbf{V}$)
matrices for each of $H$ heads: $\mathbf{Q}_h = \mathbf{Z}_X
\mathbf{W}_Q^{(h)}$, $\mathbf{K}_h = \mathbf{Z}_X \mathbf{W}_K^{(h)}$,
$\mathbf{V}_h = \mathbf{Z}_X \mathbf{W}_V^{(h)}$, where
$\mathbf{W}_Q^{(h)}, \mathbf{W}_K^{(h)}, \mathbf{W}_V^{(h)} \in
\mathbb{R}^{d_m \times d_k}$ and $d_k = d_m / H$. The (masked) attention
mechanism is defined in Equation \ref{eq:attention}
\begin{equation}
\mathrm{Attn}(\mathbf{Q}_h, \mathbf{K}_h, \mathbf{V}_h) =
\mathrm{softmax}\!\left(\frac{\mathbf{Q}_h \mathbf{K}_h^{\mathsf{T}}}{\sqrt{d_k}}\right) \mathbf{V}_h .
\label{eq:attention}
\end{equation}
The multi-head output is $\mathrm{MHSA}(\mathbf{Z}_X) =
\mathrm{Concat}(\mathrm{head}_1, \dots, \mathrm{head}_H)\mathbf{W}_O$,
with $\mathbf{W}_O \in \mathbb{R}^{d_m \times d_m}$.

\textit{2. Residual Connection and Normalization.} To facilitate stable
gradient flow, we apply layer normalization (LN) over the residual sum,
following the post-layer-normalization convention of the original
Transformer formulation: $\mathbf{Z}' = \mathrm{LN}(\mathbf{Z}_X +
\mathrm{MHSA}(\mathbf{Z}_X))$.

\textit{3. Position-wise Feed-Forward Network (FFN).} A two-layer
transformation is applied to each position as shown in Equation \ref{eq:ffn}:
\begin{equation}
\mathrm{FFN}(\mathbf{x}) = \mathrm{ReLU}(\mathbf{x} \mathbf{W}_1 + \mathbf{b}_1) \mathbf{W}_2 + \mathbf{b}_2 ,
\label{eq:ffn}
\end{equation}
where $\mathbf{W}_1 \in \mathbb{R}^{d_m \times d_{ff}}$ and
$\mathbf{W}_2 \in \mathbb{R}^{d_{ff} \times d_m}$, with $d_{ff} =
4d_m$.

\textit{4. Final Stream Representations.} The final output for each
stream $X \in \{\mathrm{Ag}, \mathrm{H}, \mathrm{L}\}$ is a refined
feature matrix $\mathbf{T}_X = \mathrm{LN}(\mathbf{Z}' +
\mathrm{FFN}(\mathbf{Z}')) \in \mathbb{R}^{L_X \times d_m}$.

\subsubsection{CNN Branch}

Operating in parallel to the Transformer branch, the CNN branch uses 1D
CNNs to capture local biophysical motifs indicative of
paratope--epitope hotspots. For a projected input sequence
$\mathbf{Z}_X \in \mathbb{R}^{L \times d_m}$, the architecture proceeds
as follows.

\textit{1. 1D Convolutional Layers.} Two sequential layers extract
hierarchical local features. The activation $O_{k,j}$ for the $k$-th
filter at position $j$ is given iin Equation \ref{eq:conv}
\begin{equation}
O_{k,j} = \mathrm{ReLU}\!\left( \sum_{i=0}^{F_W - 1} \sum_{d=0}^{C_{in} - 1}
\mathbf{W}_{k,i,d} \cdot \mathbf{Z}_X[j+i, d] + B_k \right),
\label{eq:conv}
\end{equation}
where $C_{in}$ is the input channel depth (e.g., $d_m = 256$ for the
first layer), $\mathbf{W}$ and $B$ are learnable weights and biases,
and $F_W$ is the filter width.

\textit{2. Global Feature Aggregation.} To extract sequence-invariant
descriptors, we apply global average pooling (GAP) across the temporal
(residue) dimension, restricted to non-padded positions
($j \in \{1,\dots,L\}$, Equation \ref{eq:conv_gap}):
\begin{equation}
P_k = \frac{1}{L} \sum_{j=1}^{L} O_{k,j} .
\label{eq:conv_gap}
\end{equation}

\textit{3. Stream-Specific Outputs.} This yields a fixed-size compressed
vector $\mathbf{C}_X \in \mathbb{R}^{D_{\mathrm{CNN}}}$ for each stream
$X \in \{\mathrm{Ag}, \mathrm{H}, \mathrm{L}\}$, with $D_{\mathrm{CNN}} =
128$.

\subsection{Feature Fusion and Affinity Regression}

To integrate the multi-scale representations, we employ a late-fusion
strategy intended to preserve the distinct biochemical identities of the
antigen and the antibody chains.

\textit{1. Branch-Wise Aggregation.} For each stream $X \in
\{\mathrm{Ag}, \mathrm{H}, \mathrm{L}\}$, the Transformer output
$\mathbf{T}_X \in \mathbb{R}^{L_X \times d_m}$ is condensed into a
fixed-size vector $\bar{\mathbf{T}}_X \in \mathbb{R}^{d_m}$ by GAP over
the non-padded sequence positions (Equation \ref{eq:branchpool})
\begin{equation}
\bar{\mathbf{T}}_X = \frac{1}{L_X} \sum_{t=1}^{L_X} \mathbf{T}_X[t,:] \in \mathbb{R}^{d_m},
\label{eq:branchpool}
\end{equation}
capturing the sequence-wide evolutionary context.

\textit{2. Intra-Stream Fusion.} The global context
($\bar{\mathbf{T}}_X$) and local motif ($\mathbf{C}_X$) features are
concatenated into a 384-dimensional stream-specific descriptor (Equation \ref{eq:intra_fusion}):
\begin{equation}
\mathbf{f}_X = [\bar{\mathbf{T}}_X \mathbin{\|} \mathbf{C}_X] \in \mathbb{R}^{d_m + D_{\mathrm{CNN}}}.
\label{eq:intra_fusion}
\end{equation}

\textit{3. Global Late Fusion.} The final interaction descriptor
$\mathbf{F}_{\mathrm{total}}$ concatenates the three stream vectors (Equation \ref{eq:global_fusion}):
\begin{equation}
\mathbf{F}_{\mathrm{total}} = [\mathbf{f}_{\mathrm{Ag}} \mathbin{\|} \mathbf{f}_{\mathrm{H}} \mathbin{\|} \mathbf{f}_{\mathrm{L}}] \in \mathbb{R}^{1152}.
\label{eq:global_fusion}
\end{equation}

This late-fusion design is intended to avoid the ``information
dilution'' risk of early-concatenation models by letting each branch
optimize its representation independently. $\mathbf{F}_{\mathrm{total}}$
is passed to a three-layer MLP head that performs the final non-linear
mapping to the affinity score $\hat y$ (Eq.~\ref{eq:finalpred};
architectural details in the DuaDeep Architecture Details appendix,
below). This design
balances the need for high-resolution motif extraction with the
computational efficiency required for high-throughput
paratope--epitope screening.


\section{Results and Discussion}
\label{results}

\subsection{Dataset}

The experiments were conducted on the publicly available \textit{AbRank}
dataset,\cite{Plissier2024AbRank} which provides experimentally
validated antigen--antibody pairs together with their binding affinities
(expressed as $K_d$ values in nM). Each record includes the amino acid
sequences of the antigen and the antibody heavy and light chains,
enabling sequence-level modeling of antigen--antibody interactions.

\paragraph{Preprocessing.} Raw sequences were cleaned to retain only
standard amino-acid characters (A--Y). Nonstandard or missing residues
were replaced with the placeholder token ``X''. Entries with missing or
invalid affinity values were removed, and affinity values were filtered
to a biologically meaningful range ($10^{-3} < K_d < 10^{9}$ nM). The
cleaned $K_d$ values were converted to negative logarithmic form
$\mathrm{p}K_d = 9 - \log_{10}(K_d)$, which produces a scale positively
correlated with binding strength.

\paragraph{Data Splitting.} To rigorously evaluate the generalization
capacity of DuaDeep-SeqAffinity, we implemented a sequence-disjoint
splitting protocol to ensure a strict separation of the underlying
biological entities. We first identified the complete set of unique
antibody sequences and unique antigen sequences across the dataset.
These unique sequences were then randomly assigned to training (80\%),
validation (10\%), and testing (10\%) subsets. An antibody--antigen pair
was assigned to a specific subset only if both its constituent antibody
and antigen sequences were members of that subset's unique-sequence
pool -- i.e., the ``C3'' generalization regime in the taxonomy of Park
and Marcotte, in which neither partner of a test pair has been observed
during training.\cite{park2012flaws} 

\subsection{Model Performance}

To assess the predictive robustness of DuaDeep-SeqAffinity, we
conducted a systematic ablation study comparing our dual-stream hybrid
architecture against its constituent single-branch variants: (1) ESM-T
(ESM-2 + Transformer) and (2) ESM-C (ESM-2 + CNN). All models were
evaluated across four independent runs with different random seeds
(Table~\ref{tab:hyperparams}). As summarized in Table~\ref{tab:results},
the proposed model achieves the best performance across every metric,
reaching a mean RMSE of 0.7366, a mean MAE of 0.5147, and the highest
coefficient of determination ($R^2 = 0.460$). DuaDeep-SeqAffinity also
yields the strongest alignment with experimental data among the three
architectures, attaining a mean Pearson correlation of 0.683 and a mean
Spearman coefficient of 0.683, together with the highest discriminative
power in ranking (mean Pairwise AUC of 0.895).

\begin{table*}[h]
\centering
\caption{Test-set performance across four runs with different random
seeds for each model architecture, on the affinity-regression task. The
table reports regression metrics (RMSE, MAE, and $R^2$), correlation
coefficients (Pearson and Spearman), and pairwise ranking AUC for ESM-2
combined with a Transformer, a CNN, and the dual-stream DuaDeep model.}
\label{tab:results}
\resizebox{\textwidth}{!}{%
\begin{tabular}{l l cccccc}
\toprule
Model & Run & RMSE $\downarrow$ & MAE $\downarrow$ & $R^2$ $\uparrow$ & Pearson $\uparrow$ & Spearman $\uparrow$ & Pairwise AUC $\uparrow$ \\
\midrule
\multirow{5}{*}{Transformer (ESM-T)}
 & Run 1  & 0.8369 & 0.6327 & 0.307 & 0.576 & 0.556 & 0.824 \\
 & Run 2  & 0.8094 & 0.5926 & 0.322 & 0.607 & 0.590 & 0.848 \\
 & Run 3  & 0.7581 & 0.5691 & 0.413 & 0.645 & 0.628 & 0.872 \\
 & Run 4  & 0.7985 & 0.5822 & 0.357 & 0.625 & 0.603 & 0.852 \\
\cmidrule(lr){2-8}
 & Mean $\pm$ SD & \textbf{0.8007 $\pm$ 0.033} & \textbf{0.5942 $\pm$ 0.027} & \textbf{0.350 $\pm$ 0.047} & \textbf{0.613 $\pm$ 0.030} & \textbf{0.594 $\pm$ 0.030} & \textbf{0.849 $\pm$ 0.020} \\
\midrule
\multirow{5}{*}{CNN (ESM-C)}
 & Run 1  & 0.7744 & 0.5726 & 0.395 & 0.633 & 0.611 & 0.853 \\
 & Run 2  & 0.7667 & 0.5720 & 0.392 & 0.634 & 0.608 & 0.857 \\
 & Run 3  & 0.7733 & 0.5593 & 0.390 & 0.642 & 0.621 & 0.859 \\
 & Run 4  & 0.7733 & 0.5722 & 0.397 & 0.636 & 0.614 & 0.854 \\
\cmidrule(lr){2-8}
 & Mean $\pm$ SD & \textbf{0.7719 $\pm$ 0.004} & \textbf{0.5690 $\pm$ 0.007} & \textbf{0.394 $\pm$ 0.003} & \textbf{0.636 $\pm$ 0.004} & \textbf{0.614 $\pm$ 0.006} & \textbf{0.856 $\pm$ 0.003} \\
\midrule
\multirow{5}{*}{DuaDeep}
 & Run 1  & 0.7364 & 0.5193 & 0.467 & 0.684 & 0.683 & 0.894 \\
 & Run 2  & 0.7300 & 0.5102 & 0.462 & 0.684 & 0.685 & 0.897 \\
 & Run 3  & 0.7425 & 0.5178 & 0.451 & 0.677 & 0.683 & 0.899 \\
 & Run 4  & 0.7373 & 0.5116 & 0.460 & 0.688 & 0.680 & 0.890 \\
\cmidrule(lr){2-8}
 & Mean $\pm$ SD & \textbf{0.7366 $\pm$ 0.005} & \textbf{0.5147 $\pm$ 0.005} & \textbf{0.460 $\pm$ 0.007} & \textbf{0.683 $\pm$ 0.005} & \textbf{0.683 $\pm$ 0.002} & \textbf{0.895 $\pm$ 0.004} \\
\bottomrule
\end{tabular}%
}
\end{table*}

\paragraph{Statistical Significance of the Dual-Branch Improvement.}
To assess whether the improvements in Table~\ref{tab:results} are
statistically robust rather than incidental to the specific seeds
evaluated, we performed paired, two-sided $t$-tests across the four
matched random seeds, comparing DuaDeep against each single-branch
ablation on RMSE, Pearson correlation, and Pairwise AUC
(Table~\ref{tab:sig}). Because each configuration was trained and
evaluated with the same seed set (Table~\ref{tab:hyperparams}), the
comparisons are paired at the seed level, which increases statistical
power relative to an unpaired test. 
DuaDeep significantly outperformed both
ablations on all three metrics ($p < 0.05$ in every comparison, and $p <
0.01$ in five of six), with particularly large and consistent effect
sizes relative to ESM-C (paired Cohen's $d_z > 5$ for Pearson
correlation and Pairwise AUC). These results support the claim that the
dual-branch fusion provides a systematic, rather than seed-specific,
improvement over either single-branch architecture. 

\begin{table}[h]
\centering
\caption{Statistical significance of the DuaDeep-SeqAffinity
improvement over each single-branch ablation, computed directly from the
per-seed results in Table~\ref{tab:results} (paired, two-sided
$t$-test, $n = 4$ matched seeds; $d_z$ = paired Cohen's $d$).}
\label{tab:sig}
\begin{tabular}{llrrr}
\toprule
Comparison & Metric & Mean $\Delta$ & $t(3)$ & $p$ \\
\midrule
DuaDeep vs.\ ESM-T & RMSE         & $-0.064$ & $-3.55$  & 0.038 \\
DuaDeep vs.\ ESM-C & RMSE         & $-0.035$ & $-22.39$ & $<0.001$ \\
DuaDeep vs.\ ESM-T & Pearson $r$  & $+0.070$ & $4.44$   & 0.021 \\
DuaDeep vs.\ ESM-C & Pearson $r$  & $+0.047$ & $11.69$  & 0.001 \\
DuaDeep vs.\ ESM-T & Pairwise AUC & $+0.046$ & $5.01$   & 0.015 \\
DuaDeep vs.\ ESM-C & Pairwise AUC & $+0.039$ & $35.40$  & $<0.001$ \\
\bottomrule
\end{tabular}
\end{table}

\begin{figure}[t]
     \centering
     \begin{subfigure}[b]{0.48\textwidth}
         \centering
         \includegraphics[width=\textwidth]{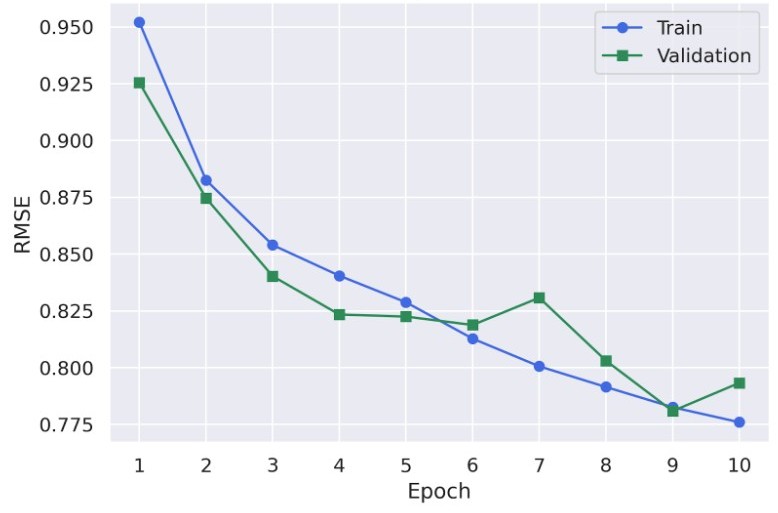}
         \caption{ESM-C: ESM-2 + CNN}
         \label{fig:esm_c}
     \end{subfigure}
     \hfill
     \begin{subfigure}[b]{0.48\textwidth}
         \centering
         \includegraphics[width=\textwidth]{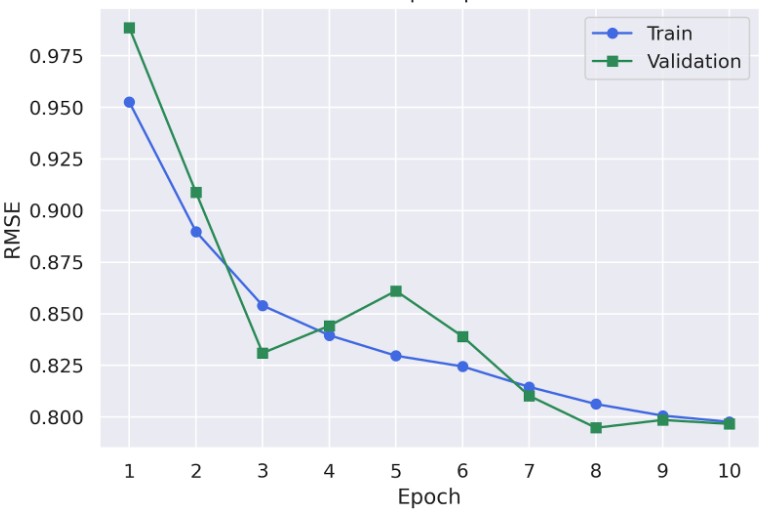}
         \caption{ESM-T: ESM-2 + Transformer}
         \label{fig:esm_t}
     \end{subfigure}

     \vspace{1em}

     \begin{subfigure}[b]{0.43\textwidth}
         \centering
         \includegraphics[width=\textwidth]{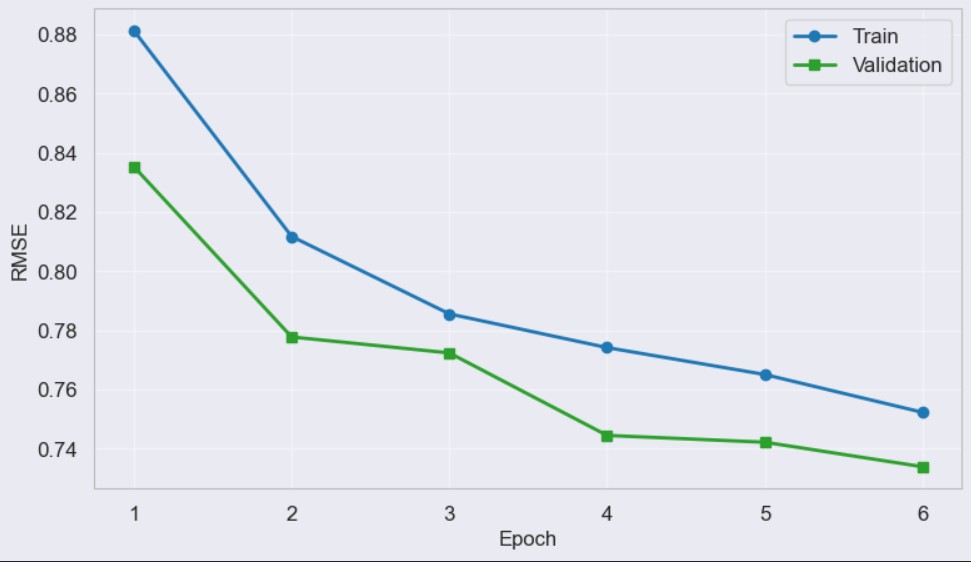}
         \caption{DuaDeep: dual-stream hybrid}
         \label{fig:duadeep_curves}
     \end{subfigure}
     
     \caption{Learning dynamics and convergence benchmarks.
     (a--c) Training and validation RMSE trajectories for the evaluated
     architectures.}
     \label{fig:combined_results}
\end{figure}

The learning dynamics across all evaluated architectures provide
further empirical evidence for the robustness of our framework.
Figure~\ref{fig:combined_results} illustrates the evolution of training
and validation RMSE over successive epochs. The ESM-C model
(Figure~\ref{fig:esm_c}) exhibits relatively slower convergence and a
persistent generalization gap, suggesting that local feature extraction
alone is insufficient for capturing the complexity of the
paratope--epitope interface. ESM-T (Figure~\ref{fig:esm_t}) shows
improved convergence relative to ESM-C but displays higher volatility in
validation loss than the hybrid approach. In contrast, the
DuaDeep-SeqAffinity model (Figure~\ref{fig:duadeep_curves}) demonstrates
closely aligned training and validation curves and the lowest recorded
validation RMSE, 0.7338, by the sixth epoch. We note that the DuaDeep
training log in Figure~\ref{fig:duadeep_curves} contains six recorded
epochs, 
this is consistent with our
best-validation-loss checkpoint selection criterion, under which
DuaDeep's optimum was reached earliest. A direct comparison of validation trajectories shows the dual-stream design
reaching a lower error floor than either single-stream baseline.

\paragraph{Comparison with the Literature.} To contextualize the
performance of \textit{DuaDeep-SeqAffinity}, we compare against several
contemporary benchmarks, including DG-Affinity,\cite{yuan2023dg}
MVSF-AB,\cite{li2025mvsf} and the structure-aware
WALLE-Affinity.\cite{liu2025abrank} We stress that, unlike the paired
ablation comparisons above, the DG-Affinity, MVSF-AB, WALLE-Affinity,
ESM-2+AntiBERTy, and Mint figures in Tables~\ref{tab:sota_extended_reg}
and~\ref{tab:sota_extended} are drawn directly from their respective
publications.\\ 
As summarized in Table~\ref{tab:sota_extended_reg}, our model achieves a
Pearson correlation of \textbf{0.683}, numerically above the centralized
backbone of DG-Affinity (0.6556), and a substantially lower RMSE
(0.737) than the value reported for MVSF-AB (1.447),  although we note that RMSE is scale-dependent and the two studies may not report
$\mathrm{p}K_d$ on an identical numerical scale. 
 In terms of
ranking discriminative power (Table~\ref{tab:sota_extended}),
\textit{DuaDeep-SeqAffinity} reaches a Pairwise AUC of \textbf{0.895},
above the sequence-only benchmarks Mint\cite{ullanat2025learning}
(0.775) and ESM-2+AntiBERTy\cite{ruffolo2021deciphering} (0.761), and
numerically matches or exceeds the ranking AUC reported for
WALLE-Affinity (0.866), a model that additionally leverages explicit
three-dimensional structural coordinates. 

\begin{table}[t]
\centering
\caption{Comparative performance of DuaDeep-SeqAffinity against
baseline architectures and literature-reported sequence-only models.
DG-Affinity and MVSF-AB values are as reported in the cited
publications.}
\label{tab:sota_extended_reg}
\begin{tabular}{lccccc}
\toprule
\textbf{Metric} & \textbf{DG-Aff.\cite{yuan2023dg}} & \textbf{MVSF-AB\cite{li2025mvsf}} & \textbf{ESM-C} & \textbf{ESM-T} & \textbf{DuaDeep} \\
\midrule
Pearson & 0.6556 & -- & 0.636 & 0.613 & \textbf{0.683} \\
$R^2$   & --     & 0.467 & 0.394 & 0.350 & \textbf{0.460} \\
RMSE    & --     & 1.447 & 0.772 & 0.80 & \textbf{0.737} \\
\bottomrule
\end{tabular}
\end{table}

\begin{table}[t]
\centering
\caption{Comparison of DuaDeep-SeqAffinity against state-of-the-art
(SOTA) methods for antigen--antibody interaction ranking. AUC values for
non-DuaDeep entries are as reported in the respective publications
(see caveat in main text), not recomputed on our split.}
\label{tab:sota_extended}
\small
\begin{tabular}{lcc}
\toprule
\textbf{Model} & \textbf{Input Type} & \textbf{Pairwise AUC} \\
\midrule
WALLE-Affinity (ranking)\cite{liu2025abrank} & Structure + sequence & 0.866 \\
WALLE-Affinity (regression)\cite{liu2025abrank} & Structure + sequence & 0.833 \\
ESM-2 + AntiBERTy (ranking)\cite{ruffolo2021deciphering} & Sequence-only & 0.761 \\
Mint (ranking)\cite{ullanat2025learning} & Sequence-only & 0.775 \\
\midrule
ESM-C (ESM-2 + CNN) & Sequence-only (ours) & 0.856 \\
ESM-T (ESM-2 + Transformer) & Sequence-only (ours) & 0.850 \\
\textbf{DuaDeep (ESM-2 + Trans.\ + CNN)} & Sequence-only (ours) & \textbf{0.895} \\
\bottomrule
\end{tabular}
\end{table}

\subsection{Model Interpretability}

\begin{figure}[tb]
    \centering
    \includegraphics[width=\linewidth]{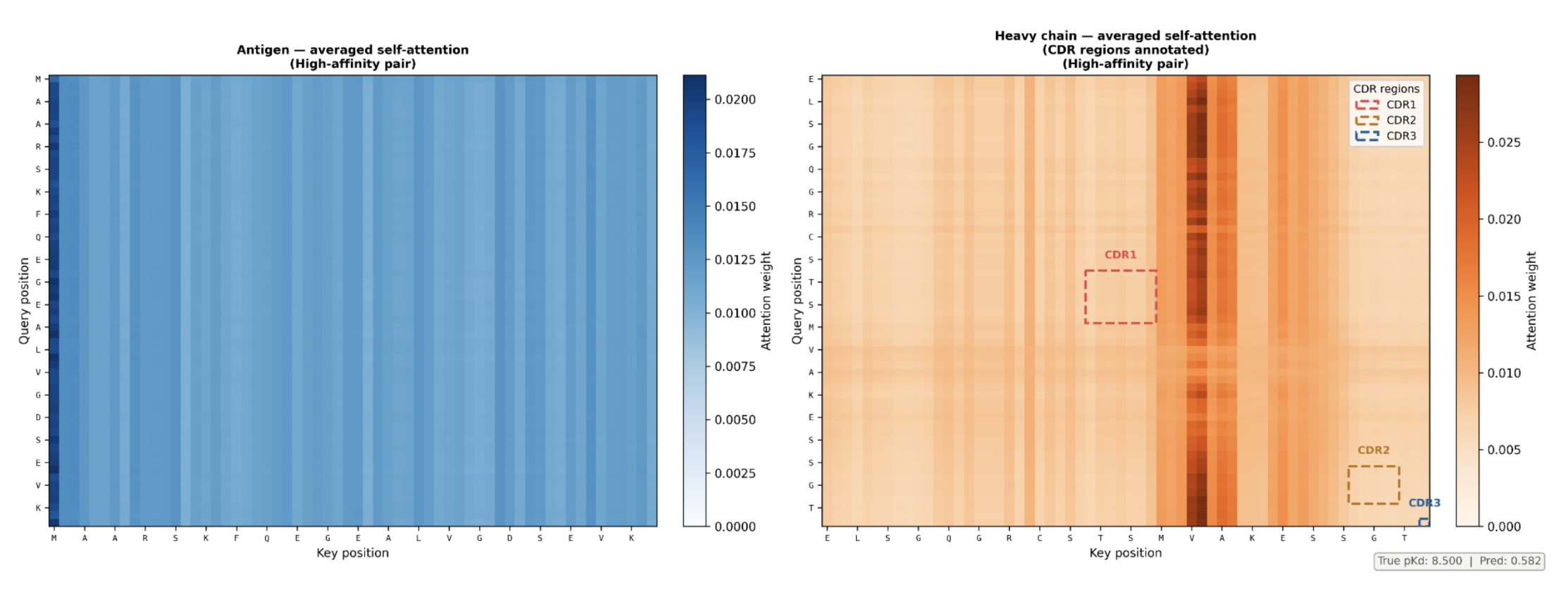}
    \caption{Interpretability analysis of the Transformer stream. Heatmaps
    show self-attention weights averaged over all heads and layers for a
    representative high-affinity pair. (Left) Antigen attention shows a
    structured focus on key residues. (Right) Antibody heavy-chain
    attention shows localized focus on the CDR regions (annotated in
    red, gold, and blue).}
    \label{fig:attention_maps}
\end{figure}

To investigate the internal representation logic of the Transformer
branch, we visualized the multi-head self-attention maps averaged
across all layers and heads for a representative high-affinity
antibody--antigen pair. As shown in Figure~\ref{fig:attention_maps}, the
model exhibits a selective attentional focus on structurally significant
regions. The antibody heavy-chain attention map (right) shows
high-weight clusters that align with the hypervariable CDR1 and CDR2
loops. On the antigen side (left), the model exhibits structured
vertical ``striping,'' identifying candidate anchor residues that serve
as focal points for the Transformer's global representation.

\begin{figure}[tb]
    \centering
    \includegraphics[width=\linewidth]{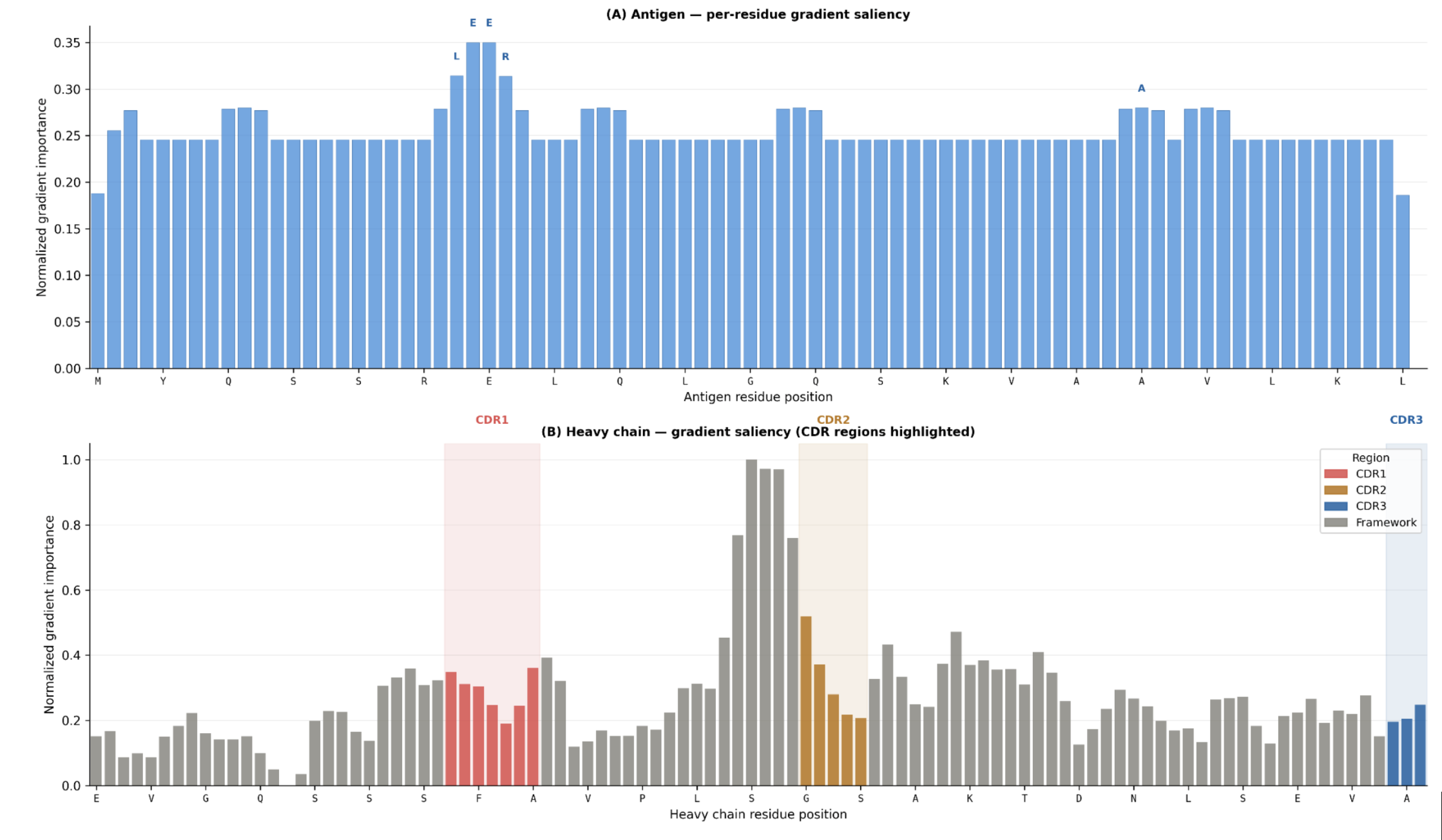}
    \caption{Gradient-based saliency maps. (A) Antigen saliency
    identifies candidate epitopic residues (labeled) with high
    sensitivity to affinity perturbation. (B) Heavy-chain saliency shows
    localized importance in CDR1 (red), CDR2 (gold), and CDR3 (blue).}
    \label{fig:saliency_maps}
\end{figure}

To further quantify the model's decision-making, we computed the
$L_2$-normalized magnitude of the gradient of the predicted affinity
$\hat y$ with respect to each input embedding (vanilla saliency
mapping). As illustrated in Figure~\ref{fig:saliency_maps}A, this
sensitivity analysis identifies candidate ``epitopic hotspots'' on the
antigen, including a high-sensitivity cluster around residues L, E, E,
and R. On the antibody heavy chain (Figure~\ref{fig:saliency_maps}B),
gradient importance is concentrated within the CDR1, CDR2, and CDR3
loops, as well as several supporting framework (FW) residues.

\subsection{Discussion and Limitations}
\label{limitatiions}

\paragraph{Architectural Synergy and Inductive Bias.} The performance
gains of \textit{DuaDeep-SeqAffinity} are consistent with a decoupled
inductive bias that enforces a ``separation of concerns'' between local
motifs and global context. By maintaining independent feature streams,
the design appears to mitigate the ``signal dilution'' common in
monolithic backbones, helping preserve high-variance local signal in
CDR loops. The stability observed in Figure~\ref{fig:combined_results},
together with the paired significance tests in Table~\ref{tab:sig},
suggests that this synergy provides a systematic, non-incidental
benefit over either constituent branch alone. DuaDeep additionally
matches or exceeds the ranking performance reported for the
structure-aware WALLE-Affinity baseline, consistent with the hypothesis
that multi-scale sequence integration can recover some of the interface
information that explicit three-dimensional modeling provides; we
present this as a promising signal.

\paragraph{Biophysical Plausibility.} Interpretability analysis via
attention (Figure~\ref{fig:attention_maps}) and saliency mapping
(Figure~\ref{fig:saliency_maps}) is consistent with the model's
predictive logic drawing on biophysically relevant regions: the
emergent focus on CDR loops and candidate epitopic anchor residues
indicates that \textit{DuaDeep} can identify functional paratopes
directly from primary sequence. This evidence is
currently qualitative and example-level; extending it to population-level,
quantitative overlap statistics,  is an important next step.

\paragraph{Limitations and Future Work.} While our late-fusion strategy
ensures scalability for high-throughput screening, the use of GAP
creates a localized information bottleneck that discards explicit
residue-to-residue spatial coordination. Future iterations will explore
cross-attention mechanisms to resolve such coordination between binding
partners explicitly. Beyond cross-attention, promising directions
include homology-aware data splitting to more rigorously bound
generalization error; systematic ablation of ESM-2 backbone scale;
decoupling the optimization schedule from architectural comparisons;
and population-level, rather than single-example, quantitative
validation of interpretability claims. Extending the framework to
broader mutational landscapes and zero-shot generalization, potentially
building on the mutation-effect sensitivity already demonstrated by
protein language models,\cite{meier2021zeroshot} remains a priority for
advancing AI-driven therapeutic antibody discovery.

\section{Conclusions}
\label{conclusion}

In this work, we presented DuaDeep-SeqAffinity, a three-stream,
dual-branch architecture for sequence-only antibody--antigen affinity
prediction. By processing the antigen and antibody chains through
parallel Transformer and CNN branches under a decoupled inductive bias,
the framework integrates global evolutionary context with local
biophysical motifs and achieves, to our knowledge, the strongest
sequence-only pairwise-ranking performance reported to date on the
AbRank benchmark (Pairwise AUC $= 0.895 \pm 0.004$), together with
consistent, statistically significant improvements over single-branch
ablations. DuaDeep-SeqAffinity also compares
favorably with several structure-aware baselines evaluated on related
benchmarks, suggesting that multi-scale sequence integration can
recover a meaningful portion of the interface information that explicit
three-dimensional modeling provides.
Attention and gradient-based saliency analyses provide qualitative
evidence that the model's decision-making concentrates on CDR loops and
candidate epitope residues, consistent with known antibody--antigen
biology, while we emphasize that this evidence should be extended to
population-level statistics in future work. With an $R^2$ of 0.46, underscoring that
DuaDeep-SeqAffinity is best positioned as a scalable pre-screening and
ranking tool. 
Future work will pursue homology-aware benchmarking, cross-attention interaction modeling to resolve explicit residue-to-residue coordination between binding partners, and evaluation across a broader range of ESM-2/ESM3 backbone
scales, to further advance the frontier of AI-driven biologics
discovery.

\appendix

\section{Reproducibility Details}
\label{app:param_details}

Table~\ref{tab:hyperparams} summarizes the key training hyperparameters
shared across the ESM-T, ESM-C, and DuaDeep model variants.

\begin{table}[h]
\centering
\caption{Training hyperparameters shared across the ESM-T, ESM-C, and
DuaDeep model variants.}
\label{tab:hyperparams}
\begin{tabular}{ll}
\toprule
\textbf{Hyperparameter} & \textbf{Value} \\
\midrule
Optimizer & AdamW \\
Learning rate & $1 \times 10^{-4}$ \\
Weight decay & $1 \times 10^{-4}$ \\
Batch size & 16 \\
Loss function & Smooth L1 (Huber), $\beta = 0.5$ \\
Mixed precision & FP16 via \texttt{torch.amp.GradScaler} \\
Max sequence length & 512 tokens \\
Random seeds & $\{5, 53, 100, 42\}$ \\
Data split & 80\% train / 10\% val / 10\% test \\
ESM-2 backbone & \texttt{facebook/esm2\_t12\_35M\_UR50D} (frozen) \\
Model selection & Best validation loss (no patience / no LR scheduler) \\
\bottomrule
\end{tabular}
\end{table}

\section{Transformer Architecture Details}

Table~\ref{tab:transformer-arch} reports the architectural
hyperparameters of the Transformer branch used in ESM-T.

\begin{table}[h]
\centering
\caption{Transformer branch architecture and training configuration.}
\label{tab:transformer-arch}
\begin{tabular}{ll}
\toprule
\textbf{Parameter} & \textbf{Value} \\
\midrule
Number of attention heads & 8 \\
Number of Transformer encoder layers (per stream) & 2 \\
Feed-forward hidden dimension & 1024 \\
Projection dimension ($d_{\mathrm{model}}$) & 256 \\
Number of training epochs & 10 \\
Learning rate schedule & Constant (no scheduler) \\
\bottomrule
\end{tabular}
\end{table}

\section{DuaDeep Architecture Details}
\label{app:duadeep_arch}

DuaDeep employs a dual-stream hybrid feature extractor that combines
both the Transformer and CNN branches within each input stream.
Table~\ref{tab:duadeep-arch} summarizes its architectural configuration.
Within each stream, the projected ESM-2 embeddings ($\mathbf{Z} \in
\mathbb{R}^{L \times 256}$) are fed to the Transformer and CNN branches
in parallel; the Transformer branch captures long-range sequential
dependencies via self-attention, while the CNN branch extracts local
motif patterns through successive convolutions with kernel sizes 3 and
5. Their pooled outputs are concatenated into a 384-dimensional
per-stream representation, which is then fused across the three
streams (antigen, heavy chain, light chain) at the prediction head,
\begin{equation}
\hat{y} = \mathrm{MLP}\!\left(\left[\mathbf{f}_{\mathrm{Ag}} \mathbin\| \mathbf{f}_{\mathrm{H}} \mathbin\| \mathbf{f}_{\mathrm{L}}\right]\right).
\label{eq:finalpred}
\end{equation}

\begin{table}[h]
\centering
\caption{DuaDeep hybrid feature-extractor architecture.}
\label{tab:duadeep-arch}
\begin{tabular}{ll}
\toprule
\textbf{Parameter} & \textbf{Value} \\
\midrule
\multicolumn{2}{l}{\textit{Transformer branch (per stream)}} \\
\quad Number of attention heads & 8 \\
\quad Number of encoder layers & 2 \\
\quad Feed-forward hidden dimension & 1024 \\
\quad Projection dimension ($d_{\mathrm{model}}$) & 256 \\
\quad Pooling & Adaptive average pool $\to$ 256-dim \\
\midrule
\multicolumn{2}{l}{\textit{CNN branch (per stream)}} \\
\quad Conv1d layer 1 & $256 \!\to\! 128$, kernel size 3, padding \texttt{same} \\
\quad Conv1d layer 2 & $128 \!\to\! 128$, kernel size 5, padding \texttt{same} \\
\quad Activation & ReLU (after each Conv1d) \\
\quad Pooling & Adaptive average pool $\to$ 128-dim \\
\midrule
\multicolumn{2}{l}{\textit{Fusion and prediction}} \\
\quad Per-stream output & $[f_{\mathrm{TF}} \mathbin\| f_{\mathrm{CNN}}] = 256 + 128 = 384$ dim \\
\quad MLP input & $384 \times 3 = 1152$ dim (3 streams concatenated) \\
\quad MLP layers & $1152 \!\to\! 1024 \!\to\! 512 \!\to\! 1$ \\
\quad MLP activation / dropout & ReLU / 0.1 (after each hidden layer) \\
\quad Number of training epochs & 10 \\
\quad Learning rate schedule & Constant (no scheduler) \\
\bottomrule
\end{tabular}
\end{table}

\section{Model Complexity Analysis}
\label{complexity}

Table~\ref{tab:params} breaks down the trainable parameters for the
ESM-T, ESM-C, and DuaDeep model variants. The frozen ESM-2 backbone
(\texttt{esm2\_t12\_35M\_UR50D}, ${\sim}35$M parameters) is excluded
from the trainable count. We independently verified every entry in
Table~\ref{tab:params} by direct calculation from the layer
dimensions in Table~\ref{tab:duadeep-arch} (e.g., the shared
projection has $480 \times 256 + 256 = 123\,136$ parameters; each
Transformer encoder layer has $789\,760$ parameters, giving $2 \times
789\,760 \times 3~\text{streams} = 4\,738\,560$ for the ESM-T feature
extractors; each CNN branch has $180\,480$ parameters, giving
$3 \times 180\,480 = 541\,440$ for ESM-C); in every case the reported
totals are internally consistent with the architectural specification,
which we view as a positive reproducibility signal for this component
of the paper.

\begin{table}[h]
\centering
\caption{Trainable parameter count by model component for ESM-T,
ESM-C, and DuaDeep. The ESM-2 backbone (${\sim}35$M parameters) is
frozen and excluded.}
\label{tab:params}
\begin{tabular}{lrrr}
\toprule
\textbf{Component} & \textbf{ESM-T} & \textbf{ESM-C} & \textbf{DuaDeep} \\
\midrule
ESM-2 projection (linear $480 \!\to\! 256$, shared) & 123\,136 & 123\,136 & 123\,136 \\
Feature extractors ($\times 3$ streams) & 4\,738\,560 & 541\,440 & 5\,280\,000 \\
Prediction head (MLP) & 1\,312\,769 & 919\,553 & 1\,705\,985 \\
\midrule
\textbf{Total trainable} & ${\sim}\mathbf{6.17}$\textbf{M} & ${\sim}\mathbf{1.58}$\textbf{M} & ${\sim}\mathbf{7.11}$\textbf{M} \\
\bottomrule
\end{tabular}
\end{table}

\paragraph{Computational Complexity.} The Transformer branch dominates
computational cost with self-attention complexity $\mathcal{O}(L^2
\cdot d)$ per layer per stream, where $L \leq 512$ and $d = 256$. With
three streams and two layers each, this yields six attention
computations per forward pass. The CNN branch is substantially cheaper,
at $\mathcal{O}(L \cdot k \cdot d)$ per convolution ($k \in \{3,5\}$).
The primary memory bottleneck is the frozen ESM-2 forward pass
(${\sim}35$M parameters resident in GPU memory) plus the self-attention
maps, which scale as $\mathcal{O}(3 \times 2 \times B \times H \times
L^2)$ for the Transformer variant, where $B$ is the batch size and $H =
8$ is the number of attention heads. DuaDeep incurs the cost of
\textbf{both} branches per stream -- six attention computations
(identical to ESM-T) plus six convolution operations (identical to
ESM-C) -- making it the most expensive variant. However, because the
CNN branch is lightweight (${\sim}180$K parameters per stream), DuaDeep
adds only ${\sim}15\%$ more trainable parameters than ESM-T alone
(7.11M vs.\ 6.17M).

\paragraph{Compute Resources.} All experiments were conducted on a
single compute node equipped with four NVIDIA H100 80 GB HBM3 GPUs and
an Intel Xeon Platinum 8481C CPU (52 cores, 2.70 GHz). Each model
variant was trained on a single GPU with FP16 mixed precision via
\texttt{torch.amp.GradScaler}. Training wall-clock time per seed was
approximately 60 min for ESM-T, 55 min for ESM-C, and 70 min for
DuaDeep, with peak GPU memory usage of ${\sim}1.6$ GB across all
variants, owing to the frozen ESM-2 backbone.

\section{Heavy- and Light-Chain Processing}

The heavy- and light-chain sequences are \textbf{not} concatenated or
merged prior to ESM-2 embedding. Instead, the architecture processes
three fully independent streams:
\begin{enumerate}
    \item \textbf{Antigen stream}: $S_{\mathrm{Ag}} \xrightarrow{\text{ESM-2}} E_{\mathrm{Ag}} \xrightarrow{\text{proj}} Z_{\mathrm{Ag}} \xrightarrow{\text{extractor}} f_{\mathrm{Ag}}$
    \item \textbf{Heavy-chain stream}: $S_H \xrightarrow{\text{ESM-2}} E_H \xrightarrow{\text{proj}} Z_H \xrightarrow{\text{extractor}} f_H$
    \item \textbf{Light-chain stream}: $S_L \xrightarrow{\text{ESM-2}} E_L \xrightarrow{\text{proj}} Z_L \xrightarrow{\text{extractor}} f_L$
\end{enumerate}
Each stream passes through the same frozen ESM-2 model (separate
forward passes), followed by a shared linear projection ($480 \to
256$), and then through its own dedicated feature extractor
(Transformer for ESM-T, CNN for ESM-C, or both in parallel for
DuaDeep). The three output vectors are concatenated only at the final
prediction head (Eq.~\ref{eq:finalpred}), where $\|$ denotes
concatenation along the feature dimension. This late-fusion strategy
preserves chain-specific representations throughout the feature
extraction pipeline, allowing the model to learn distinct patterns for
the antigen, heavy chain, and light chain before combining them for
affinity regression. For DuaDeep, each stream's feature extractor
produces the concatenation of the Transformer and CNN branch outputs
($f_{\mathrm{TF}} \mathbin\| f_{\mathrm{CNN}}$), yielding a
384-dimensional vector per stream (vs.\ 256 for ESM-T or 128 for
ESM-C).

\section*{Data availability statements}
The complete dataset used in this study is publicly available at  \\ \url{https://www.kaggle.com/datasets/aurlienplissier/abrank}.


\section*{Acknowledgement}

This work was carried out fully using the AI Datacenter of the National School of Artificial Intelligence, funded under grant number E049 24 0117 by the Algerian Ministry of Higher Education and Scientific Research. \\

\textbf{Ethics statement.} This study did not involve human
participants, animal subjects, or personally identifiable data and
therefore did not require ethics committee approval. 

\bibliographystyle{unsrt} 
\bibliography{duadeep_acs_template}

\end{document}